\documentclass{article}
\usepackage{spconf,amsmath,graphicx,hyperref}

\usepackage[utf8]{inputenc} % allow utf-8 input
\usepackage[T1]{fontenc}    % use 8-bit T1 fonts
\usepackage{url}            % simple URL typesetting
\usepackage{booktabs}       % professional-quality tables
\usepackage{amsfonts}       % blackboard math symbols
\usepackage{nicefrac}       % compact symbols for 1/2, etc.
\usepackage{microtype}      % microtypography
\usepackage{xcolor}         % colors

\usepackage{todonotes}
\usepackage{multirow}
\usepackage{amsmath}

\usepackage{algorithm}
\usepackage{algorithmic}

\usepackage[caption=false,font=footnotesize]{subfig}
\usepackage{booktabs}

\title{Graph-based 3D Human Pose Estimation using WiFi Signals}

\name{Jichao Chen$^{1}$, YangYang Qu$^{2}$, Ruibo Tang$^{3}$, Dirk Slock$^{1}$}
\address{%
$^{1}$ Communication Systems Department, EURECOM, France\\
$^{2}$ Digital Security Department, EURECOM, France\\
$^{3}$ RWTH Aachen University, Germany\\[4pt]
\mbox{\{jichao.chen, yangyang.qu, dirk.slock\}@eurecom.fr, ruibo.tang@rwth-aachen.de}
}
\begin{document}
%\ninept
%
\maketitle
\begin{abstract}

WiFi-based human pose estimation (HPE) has attracted increasing attention due to its resilience to occlusion and privacy-preserving compared to camera-based methods. However, existing WiFi-based HPE approaches often employ regression networks that directly map WiFi channel state information (CSI) to 3D joint coordinates, ignoring the inherent topological relationships among human joints. In this paper, we present GraphPose-Fi, a graph-based framework that explicitly models skeletal topology for WiFi-based 3D HPE. Our framework comprises a CNN encoder shared across antennas for subcarrier–time feature extraction, a lightweight attention module that adaptively reweights features over time and across antennas, and a graph-based regression head that combines GCN layers with self-attention to capture local topology and global dependencies. Our proposed method significantly outperforms existing methods on the MM-Fi dataset in various settings. The source code is available at: \url{https://github.com/Cirrick/GraphPose-Fi}.

\end{abstract}

\section{Introduction}\label{sec:introdcution}
Human pose estimation (HPE) is critical in a wide range of applications, including healthcare monitoring \cite{he2024expert}, augmented reality \cite{zhang2021vid2player}, and human-robot interaction \cite{gao2023parallel}. This problem has been extensively studied using cameras in computer vision (CV), spanning 2D HPE \cite{cao2017realtime}, 3D HPE \cite{zhao2022graformer, gong2023diffpose}, single-person pose estimation \cite{li2022mhformer}, and multi-person scenarios \cite{zhang2021direct, zheng2023deep}. Despite their progress, camera-based approaches face challenges in handling occlusions, low-light conditions, and privacy concerns. To address these limitations, radio-frequency (RF)-based sensing has emerged as a promising alternative, as RF signals can perceive human activities without direct visual input and are inherently robust to occlusions \cite{ma2019wifi}. Among RF signals, WiFi is particularly appealing for HPE because it is widely available, cost-effective, and energy-efficient \cite{ahmad2024wifi}. 

WiFi-based HPE leverages the variation in channel state information (CSI), which characterizes radio propagation in the surrounding environment, to infer human movements. Existing methods often treat CSI as images and apply deep regression networks to map WiFi signals into human pose coordinates. For example, WiSPPN \cite{wang2019can} employs convolutional neural networks (CNNs) to estimate the 2D pose of a single person. WiPose \cite{jiang2020towards} and GoPose \cite{ren2022gopose} extend this idea to 3D pose estimation by using CNNs to extract CSI features and long short-term memory (LSTM) networks to capture temporal dynamics, thereby producing smoother pose predictions. More recently, transformer-based methods such as MetaFi++ \cite{zhou2023metafi++} and Person-in-WiFi 3D \cite{yan2024person} have been proposed to estimate 2D and 3D poses for both single- and multi-person scenarios. Moreover, lightweight architectures such as HPE-Li \cite{d2024hpe} have been introduced to reduce computational complexity by incorporating selective kernel attention modules. 
% DT-Pose \cite{chen2025towards} tackles domain gaps and structural fidelity in WiFi-based HPE with a hybrid learning–decoding framework.

However, directly applying CNNs or transformers to map WiFi CSI into pose coordinates ignores the inherent topological relationships among human body joints. The human body can be naturally represented as a graph, where joints correspond to nodes and bones to edges. In camera-based HPE, this structural prior has been effectively modeled using graph convolutional networks (GCNs) \cite{zhao2022graformer}, showing strong performance compared to conventional approaches. Motivated by this success, we propose a graph-based architecture for 3D HPE using WiFi signals. Unlike GCN-based 2D-to-3D lifting methods in CV \cite{zheng2023deep}, which rely on off-the-shelf 2D pose estimators before applying GCNs for 3D reconstruction, WiFi-based HPE cannot directly leverage such intermediate supervision. To address this limitation, we design two modules to extract discriminative features from CSI and generate embeddings suitable for graph modeling. Specifically, we first employ a shared CNN encoder across all antennas to extract subcarrier–time features and project the subcarrier axis to a joint-aligned latent dimension. We then introduce a lightweight temporal–spatial attention module to adaptively select informative patterns along the compressed time and antenna dimensions. Finally, the resulting joint-level features are processed by a graph-based regression head, which combines GCN layers and self-attention \cite{vaswani2017attention} to capture both global dependencies and the topological structures of human joints.

\begin{figure*}[!t]  
  \centering
  \includegraphics[width=1.0\textwidth]{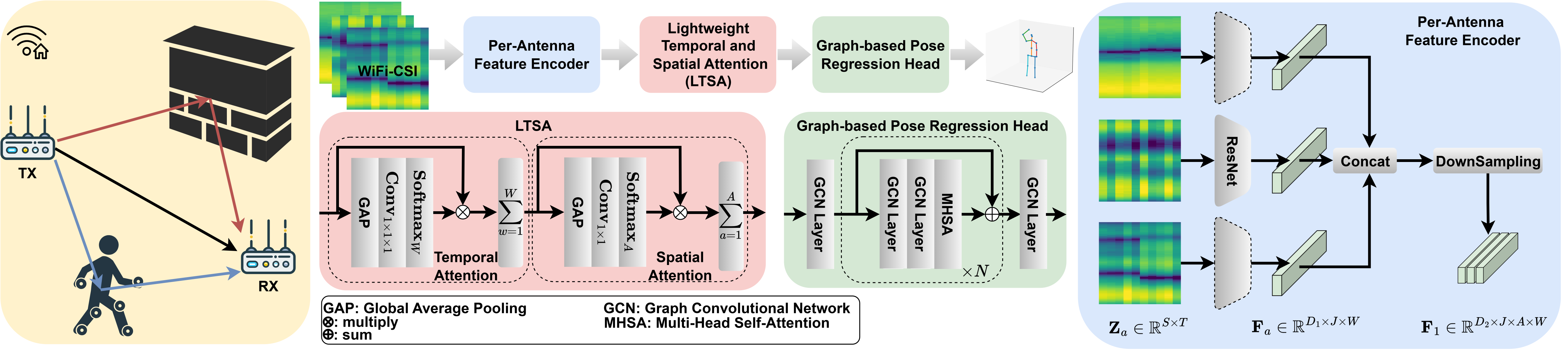}
  \caption{The overview of our proposed GraphPose-Fi for WiFi-based 3D HPE. It consists of a per-antenna feature encoder, a lightweight temporal and spatial attention module, and a graph-based pose regression head.}
  \label{fig:graphpose}
\end{figure*}

In summary, our main contributions are threefold: (i) We propose a novel graph-based framework that explicitly models the topological relationships among human joints for WiFi-based 3D HPE. (ii) We design a feature extraction strategy consisting of a CNN encoder and a lightweight temporal–spatial attention module, which produces effective embeddings for the subsequent graph-based regression head. (iii) We conduct extensive experiments on the MM-Fi public dataset \cite{yang2023mm} and our method achieves state-of-the-art results across multiple metrics and settings.
\vspace{-3mm}
\section{Method} \label{sec:method}
\vspace{-3mm}
\subsection{Preliminary}
Modern WiFi communication systems adopt orthogonal frequency division multiplexing (OFDM) \cite{5307322}, where the transmitted signals are divided into multiple subcarriers to improve spectral efficiency. Signals propagated from the transmitter (TX) to the receiver  (RX) antennas experience multipath propagation caused by reflection, diffraction, and scattering in the surrounding environment \cite{yan2024person}. These propagation effects are characterized in the CSI, which records the frequency response of each subcarrier. The movements of the human body alter the multipath components, thereby inducing distinctive variations in the CSI measurements. This is the underlying principle of WiFi-based HPE.

Given a sequence of CSI snapshots \(\mathbf{H}\in\mathbb{C}^{N_r\times N_t\times S\times T}\), where $N_r$ and $N_t$ denote the number of receive and transmit antennas, \(S\) is the number of subcarriers, and \(T\) is the number of CSI samples within a sliding temporal window aligned with one pose label, the goal is to estimate the corresponding 3D human pose coordinates \(\mathbf{Y}\in\mathbb{R}^{J\times 3}\), where \(J\) represents the number of body joints.
\vspace{-3mm}
\subsection{Proposed Method: GraphPose-Fi}
We first apply standard pre-processing such as magnitude extraction and phase calibration to convert complex CSI into a real-valued tensor \(\mathbf{Z}\in\mathbb{R}^{A\times S\times T}\), where \(A = N_r\times N_t\) is the number of antennas. We then propose GraphPose-Fi, a graph-based framework for 3D HPE using WiFi CSI, as shown in Fig.~\ref{fig:graphpose}. It consists of three modules: (1) Per-antenna feature encoder, (2) Lightweight temporal and spatial attention (LTSA) module, and (3) Graph-based pose regression head. 
\vspace{-4mm}
\subsubsection{Per-Antenna Feature Encoder}
Each antenna observes channel frequency responses under different multipath conditions, providing a complementary spatial view of human pose. Inspired by \cite{zhou2023metafi++}, we adopt a convolutional encoder based on ResNet \cite{he2016deep} shared across antennas. For antenna \(a\in\{1,\dots,A\}\), we treat its CSI slice \(\mathbf{Z}_a\in\mathbb{R}^{S\times T}\) as a two-dimensional tensor and use the encoder to obtain the features
\begin{equation}
\mathbf{F}_a \;=\; f_{\theta}(\mathbf{Z}_a), \quad
\mathbf{F}_a \in \mathbb{R}^{D_1 \times J \times W},
\end{equation}
where \(f_{\theta}\) denotes the encoder with parameters \(\theta\), \(D_1\) is the feature dimension, \(J\) is the number of latent spatial positions after convolution along subcarriers which is set to match the number of body joints for the subsequent graph modeling, and \(W\) is the compressed temporal length. We then stack per-antenna features \(\{\mathbf{F}_a\}_{a=1}^A\) and apply a point-wise convolution to reduce the feature dimension, obtaining $\mathbf{F}_1\in\mathbb{R}^{D_2\times J\times A\times W}.$
\vspace{-4mm}
\subsubsection{Lightweight Temporal and Spatial Attention}
Different time segments and antennas contribute unequally to pose estimation. We therefore design a lightweight attention module that adaptively reweights features along the temporal and spatial (antenna) dimensions, respectively. 
\par\smallskip
\noindent\textbf{Temporal Attention.}
For each joint $j$ and antenna $a$, we compute attention weights over the compressed temporal index $w \in \{1,\dots,W\}$ by first averaging along the feature dimension $D_2$, then applying a $1{\times}1{\times}1$ point-wise convolution, followed by a Softmax along the temporal axis
\begin{equation}
\alpha_{j,a,w} \;=\; \mathrm{Softmax}_{W}\!\left(\mathrm{Conv}_{1\times1\times1}\big(\mathrm{Mean}_{D_2}(\mathbf{F}_1)\big)\right).
\end{equation}
These weights are used to temporally aggregate features 
\begin{equation}
\mathbf{F}_t(:,j,a) \;=\; \sum_{w=1}^{W}\alpha_{j,a,w}\,\mathbf{F}_1(:,j,a,w), 
\quad \mathbf{F}_t \in \mathbb{R}^{D_2\times J\times A}.
\end{equation}

\noindent\textbf{Spatial Attention.}
Similarly, we derive attention weights across antennas for each joint by averaging over the feature dimension, and then applying a \(1{\times}1\) point-wise convolution followed by a Softmax along the antenna axis
\begin{equation}
\beta_{j,a} \;=\; \mathrm{Softmax}_{A}\!\left(\mathrm{Conv}_{1\times1}\big(\mathrm{Mean}_{D_2}(\mathbf{F}_t)\big)\right).
\end{equation}
The features are then aggregated across antennas
\begin{equation}
\mathbf{F}_2(:,j) \;=\; \sum_{a=1}^{A}\beta_{j,a}\,\mathbf{F}_t(:,j,a),
\quad \mathbf{F}_2 \in \mathbb{R}^{D_2\times J}.
\end{equation}
Compared to using multi-head self-attention (MHSA) \cite{vaswani2017attention} for aggregating spatiotemporal information at each joint, our method is more lightweight while achieving comparable performance, as demonstrated in the ablation study (Sec. \ref{Ablation}). Finally, we transpose $\mathbf{F}_2$ to a joint-first representation and apply LayerNorm to form the input embedding \(\mathbf{F}_3\in\mathbb{R}^{J\times D_2}\) for the next stage. 
\vspace{-3mm}
\subsubsection{Graph-based Pose Regression}
While the previous stages extract informative features across subcarriers, time, and antennas, these features are not yet explicitly aligned with the human skeleton. Motivated by graph-based HPE in CV \cite{zhao2022graformer, gong2023diffpose}, we adopt a GCN-based attention architecture that combines Chebyshev graph convolutions (ChebGConv) \cite{he2022convolutional} with self-attention to model both local topology and global dependencies.

We first map the joint-wise features $\mathbf{F}_3$ to a latent embedding \(\mathbf{X}\in\mathbb{R}^{J\times D_3}\) using a ChebGConv layer. Let $\mathbf{A}$ denote the skeleton adjacency matrix, $\mathbf{D}$ its degree matrix, $\mathbf{L} = \mathbf{I}-\mathbf{D}^{-1/2}\mathbf{A}\mathbf{D}^{-1/2}$ the normalized Laplacian and $\mathbf{I}$ the identity matrix. With the rescaled Laplacian $\tilde{\mathbf{L}} = 2\mathbf{L}/\lambda_{\max}-\mathbf{I}$, the Chebyshev convolution \cite{he2022convolutional} updates features as
\begin{equation}
\mathbf{X}^{\,l+1} \;=\; \sum_{k=0}^{K-1} \mathbf{T}_k(\tilde{\mathbf{L}})\,\mathbf{X}^{\,l}\,\mathbf{\Theta}_k,
\end{equation}
where $\mathbf{T}_k(\mathbf {\tilde{L}}) = 2{\tilde{\mathbf{L}}}\mathbf{T}_{k-1}(\mathbf {\tilde{L}})-\mathbf{T}_{k-2}(\mathbf {\tilde{L}})$ is the Chebyshev polynomial of degree \(k\) (\(\mathbf{T}_0=\mathbf{I},\,\mathbf{T}_1=\tilde{\mathbf{L}}\)), \(\lambda_{\max}\) is the largest eigenvalue of \(\mathbf{L}\), and \(\mathbf{\Theta}_k\) are learnable parameters. This expands the receptive field to \(K\)-hop neighbors.

Each GCN-based attention block contains two ChebGConv layers and an MHSA layer to aggregate global context across all joints. After $N$ such blocks, the features are mapped by a final ChebGConv layer to the predicted pose $\hat{\mathbf{Y}} \in \mathbb{R}^{J\times 3}$.
\vspace{-3mm}
\subsection{Training Objective}
We train the networks using the mean squared error (MSE) between the predicted and ground-truth joint coordinates
\begin{equation}
\mathcal{L} \;=\; \frac{1}{J}\sum_{j=1}^{J}\left\lVert \hat{\mathbf{Y}}_{j}-\mathbf{Y}_{j}\right\rVert_2^{2}.
\end{equation}

\section{Experiments}\label{sec:experiments}

\subsection{Experimental Setup}
\noindent\textbf{Dataset.} We evaluate our approach on Protocol 1 (P1) of the MM-Fi dataset~\cite{yang2023mm}, which comprises 14 daily activities performed by 40 subjects across four environments. The WiFi system operates at 5 GHz with a 40 MHz bandwidth, consisting of one transmitter with a single antenna and one receiver with three antennas. CSI is collected over 114 subcarriers, and consecutive measurements are aggregated into a sample of size $3 \times 114 \times 10$. Each pose annotation contains 17 joints in 3D coordinates. To evaluate generalization, we follow three data split strategies: (i) Random split (S1), where data are randomly divided into training and testing sets with a 3:1 ratio; (ii) Cross-subject split (S2), where 32 subjects are used for training and the remaining 8 subjects for testing; and (iii) Cross-environment split (S3), where data from three environments are used for training and one for testing.
\par\smallskip
\noindent\textbf{Evaluation Metrics.} We adopt three commonly used metrics for 3D HPE \cite{ionescu2013human3}, \cite{zheng2023deep}. Mean per joint position error (MPJPE (mm)) measures the average Euclidean distance between predicted and ground-truth joint coordinates. Procrustes aligned MPJPE (PA-MPJPE (mm)) applies a similarity transformation (translation, rotation, and scaling) before computing the error, thus focusing on the structural accuracy of poses. Percentage of correct keypoints (PCK) reports the proportion of joints whose prediction error falls within a specified threshold relative to body size, reflecting the accuracy at the joint level.
\par\smallskip
\noindent\textbf{Benchmarks.} We compare the performance of our approach with several SOTA methods, including MetaFi++ \cite{zhou2023metafi++} used as the baseline solution in MM-Fi dataset, HPE-Li \cite{d2024hpe}, and DT-Pose \cite{chen2025towards}, which are most recent and shown to achieve the best performance in MM-Fi dataset.
\par\smallskip
\noindent\textbf{Implementation Details.} We train our model using the AdamW optimizer \cite{loshchilov2017decoupled} with an initial learning rate $3\times10^{-4}$ and a weight decay 0.02. The learning rate follows a cosine decay to near zero over 50 epochs, and the batch size is 256. All experiments run on a NVIDIA GH200 GPU using PyTorch.
\vspace{-5mm}
\subsection{Comparison with State-of-the-art Methods}
Table~\ref{tab:sota_comparison} summarizes the performance of different methods on MM-Fi (Protocol 1). Our approach consistently outperforms existing baselines across all three data split settings. Under the random split (S1), it achieves the best PCK scores at all thresholds and reduces MPJPE to 160.6 mm, surpassing all the baselines. In the more challenging cross-subject split (S2), our method again achieves the highest accuracy on most metrics. For the cross-environment split (S3), where domain shift is most severe, our framework outperforms all compared methods by a clear margin across all metrics. These results demonstrate both the effectiveness and robustness of the proposed method across different experimental settings.

\begin{table}[t]
\centering
\caption{State-of-the-art performance comparisons on MM-Fi dataset (Protocol~1). $\uparrow$ higher is better; $\downarrow$ lower is better. Best values are bolded.}
\par\smallskip
\label{tab:sota_comparison}
\setlength{\tabcolsep}{4pt}\scriptsize
\begin{tabular}{@{}lccccccc@{}}
\toprule
Method &
PCK@10$\uparrow$ & 20$\uparrow$ & 30$\uparrow$ & 40$\uparrow$ & 50$\uparrow$ &
MPJPE$\downarrow$ & PA-MPJPE$\downarrow$ \\
\midrule
\multicolumn{8}{@{}l}{\textit{Setting 1 (Random Split):}}\\
MetaFi++~\cite{zhou2023metafi++} & 24.7  & 56.1 & 72.9 & 82.5 & 88.1 & 174.5    & 112.9 \\
HPE-Li~\cite{d2024hpe}           & 26.4 & 56.4 & 72.6 & 82.0 & 87.8 & 172.6    & \textbf{102.0} \\
DT-Pose~\cite{chen2025towards}   & 26.3 & 57.2 & 74.1 & 83.7 & 88.9 & 168.0    & 102.4 \\
\textbf{Ours}                    & \textbf{33.3} & \textbf{61.1} & \textbf{75.6} & \textbf{83.9} & \textbf{89.3} & \textbf{160.6} & 105.0 \\
\addlinespace[2pt]\midrule
\multicolumn{8}{@{}l}{\textit{Setting 2 (Cross-Subject):}}\\
MetaFi++~\cite{zhou2023metafi++} & 9.9 & 40.3 & 64.7 & \textbf{79.0} & \textbf{86.9} & 214.8 & 118.8 \\
HPE-Li~\cite{d2024hpe}           & 11.1 & 40.4 & 62.6 & 75.9 & 84.3 & 221.4 & \textbf{104.4} \\
DT-Pose~\cite{chen2025towards}   & 11.6 & 40.2 & 62.1 & 76.1 & 84.8 & 221.1 & 105.8 \\
\textbf{Ours}                    & \textbf{13.1} & \textbf{44.2} & \textbf{66.4} & 78.8 & 86.3 & \textbf{210.5} & 105.5 \\
\addlinespace[2pt]\midrule
\multicolumn{8}{@{}l}{\textit{Setting 3 (Cross-Environment):}}\\
MetaFi++~\cite{zhou2023metafi++} & 1.4 & 9.6 & 23.1 & 40.5 & 57.3 & 341.8 & 108.8 \\
HPE-Li~\cite{d2024hpe}           & 0.5 & 5.8 & 18.1 & 35.6 & 52.3 & 361.1 & 104.4 \\
DT-Pose~\cite{chen2025towards}   & 0.8 & 7.9 & 23.7 & 43.5 & 61.0 & 326.9 & 104.7 \\
\textbf{Ours}                    & \textbf{2.7} & \textbf{12.9} & \textbf{29.2} & \textbf{49.6} & \textbf{67.2} & \textbf{302.7} & \textbf{103.0} \\
\bottomrule
\end{tabular}
\end{table}

Table~\ref{tab:perjoint_mpjpe} further reports per-joint MPJPE on MM-Fi (P1–S1). 
Our approach achieves significant improvements on most joints, particularly in torso and head regions (e.g., bot torso, upper torso, neck base), where errors are reduced by 10–20 mm compared to the baselines. This demonstrates the effectiveness of GCN-based attention in modeling global skeletal topology. Nevertheless, all methods still exhibit higher errors on hands and elbows, which is likely due to the limited spatial resolution of WiFi signals \cite{chen2025towards}. 

\begin{table}[t]
  \centering
  \vspace{-5mm}
  \caption{Per-joint MPJPE (mm) $\downarrow$ comparisons on MM-Fi (P1--S1).}
  \label{tab:perjoint_mpjpe}
  \setlength{\tabcolsep}{5pt}
  \renewcommand{\arraystretch}{1.1}
  \scriptsize
  \begin{tabular}{lcccc}
    \toprule
    Joint & MetaFi++ \cite{zhou2023metafi++} & HPE-Li \cite{d2024hpe} & DT-Pose \cite{chen2025towards} & \textbf{Ours} \\
    \midrule
    Bot Torso    & 116.4 & 108.3 & 105.8 & \textbf{93.9} \\
    L.Hip        & 119.8 & 111.6 & 109.5 & \textbf{100.2} \\
    L.Knee       & 114.3 & 111.1 & 111.1 & \textbf{99.7} \\
    L.Foot       & 112.8 & 109.6 & 110.3 & \textbf{102.3} \\
    R.Hip        & 116.8 & 114.7 & 112.5 & \textbf{101.2} \\
    R.Knee       & 111.9 & 113.1 & 113.6 & \textbf{99.6} \\
    R.Foot       & 115.3 & 116.2 & 119.4 & \textbf{107.0} \\
    Center Torso & 117.4 & 117.7 & 116.3 & \textbf{100.4} \\
    Upper Torso  & 139.7 & 142.0 & 141.5 & \textbf{123.7} \\
    Neck Base    & 166.1 & 165.6 & 163.7 & \textbf{150.9} \\
    Center Head  & 168.9 & 166.0 & 167.6 & \textbf{150.8} \\
    R.Shoulder   & 154.5 & 153.4 & 153.4 & \textbf{141.0} \\
    R.Elbow      & 257.8 & 260.7 & \textbf{246.0} & 251.7 \\
    R.Hand       & 372.7 & 381.2 & \textbf{359.7} & 360.4 \\
    L.Shoulder   & 151.9 & 148.3 & 149.4 & \textbf{137.4} \\
    L.Elbow      & 252.6 & 244.0 & \textbf{230.4} & 244.6 \\
    L.Hand       & 377.6 & 371.2 & \textbf{346.4} & 365.2 \\
    \midrule
    Average      & 174.5 & 172.6 & 168.0 & \textbf{160.6} \\
    \bottomrule
  \end{tabular}
\end{table}
\vspace{-3mm}
\subsection{Ablation Study} \label{Ablation}
\noindent\textbf{Impact of LTSA Module.}
We investigate different strategies to aggregate the temporal and spatial information while fixing the encoder and the graph regression head. Specifically, we compare: (i) Global average pooling (GAP) over the temporal and antenna dimensions, and (ii) Per-joint multi-head self-attention (PJ-MHSA): we treat the antenna-time dimensions as tokens with features at the first dimension as embeddings. For each joint, MHSA is applied to capture intra-temporal and inter-antenna dependencies, followed by token pooling to obtain the joint embeddings. Results in Table~\ref{tab:ltsa} show that our proposed LTSA module outperforms the simple GAP baseline and achieves slightly better accuracy than PJ-MHSA while requiring substantially lower computational cost.
\par\smallskip
\noindent\textbf{Graph-Structure Design.}
To demonstrate the effectiveness of the graph-based method and test its configuration, we replace the graph regression head with a multi-layer perceptron (MLP) and vary the number of GCN-based attention blocks ($N$ from 2 to 6), while keeping the encoder and LTSA module fixed. Results in Table~\ref{tab:graph} show that our graph-based architecture significantly outperforms the MLP baseline. Moreover, performance improves when increasing the number of blocks from 2 to 4, but further stacking does not yield additional gains. Consequently, we adopt 4 GCN-based attention blocks in our model, which are effective and efficient to model the human joint topology.
\begin{table}[h]
    \centering
    \vspace{-5mm}
    \caption{Ablation study for LTSA module design.} 
    \setlength{\tabcolsep}{8pt}
    \begin{tabular}{lc}
    \toprule
    Method & MPJPE (mm) \\
    \midrule
    GAP & 161.6 \\
    PJ-MHSA        & 160.8 \\
    LTSA (ours) & \textbf{160.6} \\
    \bottomrule
    \end{tabular}
    \label{tab:ltsa}
\end{table}

\begin{table}[h]
    \centering
    \vspace{-8mm}
    \caption{Ablation study for graph-structure design.}
    \setlength{\tabcolsep}{8pt}
    \begin{tabular}{l c}
    \toprule
    Method & MPJPE (mm) \\
    \midrule
    MLP regression head & 167.8 \\
    Graph-based regression head ($N$ = 2)        & 161.8 \\
    Graph-based regression head ($N$ = 4)  & \textbf{160.6} \\
    Graph-based regression head ($N$ = 6)        & 161.5 \\
    \bottomrule
    \end{tabular} 
    \label{tab:graph}
\end{table}

\vspace{-5mm}
\section{Conclusion}\label{sec:conclusion}
We propose GraphPose-Fi, a graph-based framework for WiFi-based 3D HPE. The model couples a CNN encoder shared by antennas for subcarrier–time feature extraction with a lightweight temporal–spatial attention module, and a GCN-based attention regression head that encodes skeletal topology while capturing global dependencies. Experimental results on the MM-Fi dataset demonstrate that our approach outperforms existing state-of-the-art methods, achieving significant improvements in key metrics such as MPJPE and PCK. In future work, we will extend the framework to multi-sensor fusion, such as WiFi with RGB or radar, and investigate more robust domain generalization ability across subjects and environments.
% \par\smallskip

\noindent\textbf{Acknowledgements.} EURECOM’s research is partially supported by its industrial members: ORANGE, BMW, SAP, iABG, Norton LifeLock, by the Franco-German projects CellFree6G and 5G-OPERA, by the EU H2030 project CONVERGE, and by a Huawei France funded Chair towards Future Wireless Networks.

\clearpage
\begin{small}
    \bibliographystyle{IEEEbib}
    \bibliography{refs.bib}

@article{chen2025towards,
  title={Towards Robust and Realistic Human Pose Estimation via WiFi Signals},
  author={Chen, Yang and Guo, Jingcai and Guo, Song and Zhou, Jingren and Tao, Dacheng},
  journal={arXiv preprint arXiv:2501.09411},
  year={2025}
}

@article{zhou2023metafi++,
  title={Metafi++: Wifi-enabled transformer-based human pose estimation for metaverse avatar simulation},
  author={Zhou, Yunjiao and Huang, He and Yuan, Shenghai and Zou, Han and Xie, Lihua and Yang, Jianfei},
  journal={IEEE Internet of Things Journal},
  volume={10},
  number={16},
  pages={14128--14136},
  year={2023},
  publisher={IEEE}
}

@inproceedings{d2024hpe,
  title={HPE-Li: WiFi-Enabled Lightweight Dual Selective Kernel Convolution for Human Pose Estimation},
  author={D. Gian, Toan and Dac Lai, Tien and Van Luong, Thien and Wong, Kok-Seng and Nguyen, Van-Dinh},
  booktitle={European Conference on Computer Vision},
  pages={93--111},
  year={2024},
  organization={Springer}
}

@article{yang2023mm,
  title={Mm-fi: Multi-modal non-intrusive 4d human dataset for versatile wireless sensing},
  author={Yang, Jianfei and Huang, He and Zhou, Yunjiao and Chen, Xinyan and Xu, Yuecong and Yuan, Shenghai and Zou, Han and Lu, Chris Xiaoxuan and Xie, Lihua},
  journal={Advances in Neural Information Processing Systems},
  volume={36},
  pages={18756--18768},
  year={2023}
}

@article{ma2019wifi,
  title={WiFi sensing with channel state information: A survey},
  author={Ma, Yongsen and Zhou, Gang and Wang, Shuangquan},
  journal={ACM Computing Surveys (CSUR)},
  volume={52},
  number={3},
  pages={1--36},
  year={2019},
  publisher={ACM New York, NY, USA}
}

@inproceedings{he2016deep,
  title={Deep residual learning for image recognition},
  author={He, Kaiming and Zhang, Xiangyu and Ren, Shaoqing and Sun, Jian},
  booktitle={Proceedings of the IEEE conference on computer vision and pattern recognition},
  pages={770--778},
  year={2016}
}

@inproceedings{zhao2022graformer,
  title={Graformer: Graph-oriented transformer for 3d pose estimation},
  author={Zhao, Weixi and Wang, Weiqiang and Tian, Yunjie},
  booktitle={Proceedings of the IEEE/CVF conference on computer vision and pattern recognition},
  pages={20438--20447},
  year={2022}
}

@inproceedings{yan2024person,
  title={Person-in-wifi 3d: End-to-end multi-person 3d pose estimation with wi-fi},
  author={Yan, Kangwei and Wang, Fei and Qian, Bo and Ding, Han and Han, Jinsong and Wei, Xing},
  booktitle={Proceedings of the IEEE/CVF Conference on Computer Vision and Pattern Recognition},
  pages={969--978},
  year={2024}
}

@inproceedings{gong2023diffpose,
  title={Diffpose: Toward more reliable 3d pose estimation},
  author={Gong, Jia and Foo, Lin Geng and Fan, Zhipeng and Ke, Qiuhong and Rahmani, Hossein and Liu, Jun},
  booktitle={Proceedings of the IEEE/CVF Conference on Computer Vision and Pattern Recognition},
  pages={13041--13051},
  year={2023}
}

@article{vaswani2017attention,
  title={Attention is all you need},
  author={Vaswani, Ashish and Shazeer, Noam and Parmar, Niki and Uszkoreit, Jakob and Jones, Llion and Gomez, Aidan N and Kaiser, {\L}ukasz and Polosukhin, Illia},
  journal={Advances in neural information processing systems},
  volume={30},
  year={2017}
}

@article{loshchilov2017decoupled,
  title={Decoupled weight decay regularization},
  author={Loshchilov, Ilya and Hutter, Frank},
  journal={arXiv preprint arXiv:1711.05101},
  year={2017}
}

@article{ionescu2013human3,
  title={Human3. 6m: Large scale datasets and predictive methods for 3d human sensing in natural environments},
  author={Ionescu, Catalin and Papava, Dragos and Olaru, Vlad and Sminchisescu, Cristian},
  journal={IEEE transactions on pattern analysis and machine intelligence},
  volume={36},
  number={7},
  pages={1325--1339},
  year={2013},
  publisher={IEEE}
}

@article{he2024expert,
  title={An expert-knowledge-based graph convolutional network for skeleton-based physical rehabilitation exercises assessment},
  author={He, Tian and Chen, Yang and Wang, Ling and Cheng, Hong},
  journal={IEEE Transactions on Neural Systems and Rehabilitation Engineering},
  volume={32},
  pages={1916--1925},
  year={2024},
  publisher={IEEE}
}

@article{zheng2023deep,
  title={Deep learning-based human pose estimation: A survey},
  author={Zheng, Ce and Wu, Wenhan and Chen, Chen and Yang, Taojiannan and Zhu, Sijie and Shen, Ju and Kehtarnavaz, Nasser and Shah, Mubarak},
  journal={ACM computing surveys},
  volume={56},
  number={1},
  pages={1--37},
  year={2023},
  publisher={ACM New York, NY}
}

@article{zhang2021vid2player,
  title={Vid2player: Controllable video sprites that behave and appear like professional tennis players},
  author={Zhang, Haotian and Sciutto, Cristobal and Agrawala, Maneesh and Fatahalian, Kayvon},
  journal={ACM Transactions on Graphics (TOG)},
  volume={40},
  number={3},
  pages={1--16},
  year={2021},
  publisher={ACM New York, NY}
}

@article{gao2023parallel,
  title={Parallel dual-hand detection by using hand and body features for robot teleoperation},
  author={Gao, Qing and Ju, Zhaojie and Chen, Yongquan and Wang, Qiwen and Zhao, Yinan and Lai, Shiwu},
  journal={IEEE Transactions on Human-Machine Systems},
  volume={53},
  number={2},
  pages={417--426},
  year={2023},
  publisher={IEEE}
}

@inproceedings{cao2017realtime,
  title={Realtime multi-person 2d pose estimation using part affinity fields},
  author={Cao, Zhe and Simon, Tomas and Wei, Shih-En and Sheikh, Yaser},
  booktitle={Proceedings of the IEEE conference on computer vision and pattern recognition},
  pages={7291--7299},
  year={2017}
}

@inproceedings{li2022mhformer,
  title={Mhformer: Multi-hypothesis transformer for 3d human pose estimation},
  author={Li, Wenhao and Liu, Hong and Tang, Hao and Wang, Pichao and Van Gool, Luc},
  booktitle={Proceedings of the IEEE/CVF conference on computer vision and pattern recognition},
  pages={13147--13156},
  year={2022}
}

@article{zhang2021direct,
  title={Direct multi-view multi-person 3d pose estimation},
  author={Zhang, Jianfeng and Cai, Yujun and Yan, Shuicheng and Feng, Jiashi and others},
  journal={Advances in Neural Information Processing Systems},
  volume={34},
  pages={13153--13164},
  year={2021}
}

@article{ahmad2024wifi,
  title={WiFi-based human sensing with deep learning: Recent advances, challenges, and opportunities},
  author={Ahmad, Iftikhar and Ullah, Arif and Choi, Wooyeol},
  journal={IEEE Open Journal of the Communications Society},
  volume={5},
  pages={3595--3623},
  year={2024},
  publisher={IEEE}
}

@inproceedings{jiang2020towards,
  title={Towards 3D human pose construction using WiFi},
  author={Jiang, Wenjun and Xue, Hongfei and Miao, Chenglin and Wang, Shiyang and Lin, Sen and Tian, Chong and Murali, Srinivasan and Hu, Haochen and Sun, Zhi and Su, Lu},
  booktitle={Proceedings of the 26th Annual International Conference on Mobile Computing and Networking},
  pages={1--14},
  year={2020}
}

@article{wang2019can,
  title={Can WiFi estimate person pose?},
  author={Wang, Fei and Panev, Stanislav and Dai, Ziyi and Han, Jinsong and Huang, Dong},
  journal={arXiv preprint arXiv:1904.00277},
  year={2019}
}

@article{ren2022gopose,
  title={GoPose: 3D human pose estimation using WiFi},
  author={Ren, Yili and Wang, Zi and Wang, Yichao and Tan, Sheng and Chen, Yingying and Yang, Jie},
  journal={Proceedings of the ACM on Interactive, Mobile, Wearable and Ubiquitous Technologies},
  volume={6},
  number={2},
  pages={1--25},
  year={2022},
  publisher={ACM New York, NY, USA}
}

@ARTICLE{5307322,
  author={},
  journal={IEEE Std 802.11n-2009 (Amendment to IEEE Std 802.11-2007 as amended by IEEE Std 802.11k-2008, IEEE Std 802.11r-2008, IEEE Std 802.11y-2008, and IEEE Std 802.11w-2009)}, 
  title={IEEE Standard for Information technology-- Local and metropolitan area networks-- Specific requirements-- Part 11: Wireless LAN Medium Access Control (MAC)and Physical Layer (PHY) Specifications Amendment 5: Enhancements for Higher Throughput}, 
  year={2009},
  volume={},
  number={},
  pages={1-565},
  doi={10.1109/IEEESTD.2009.5307322}}

@article{he2022convolutional,
  title={Convolutional neural networks on graphs with chebyshev approximation, revisited},
  author={He, Mingguo and Wei, Zhewei and Wen, Ji-Rong},
  journal={Advances in neural information processing systems},
  volume={35},
  pages={7264--7276},
  year={2022}
}
\end{small}

\end{document}